\documentclass[12pt]{spieman}  
\usepackage[]{graphicx}
\usepackage{setspace}
\usepackage{tocloft}
\usepackage{threeparttable}

\title{MCL-3D: a database for stereoscopic image quality
assessment using 2D-image-plus-depth source}

\author{Rui Song,\supscr{a,b} Hyunsuk Ko,\supscr{b} C.C.Jay Kuo\supscr{b}}

\affiliation{\supscrsm{a}Xidian University, State Key Laboratory of ISN, P.O. Box 103, 2nd Taibai South Road, Xi'an, Shaanxi, China, 710071 \\
\supscrsm{b}University of Southern
California, Ming Hsieh Department of Electrical Engineering, 3740 McClintock Avenue,
Los Angeles, CA, United States, CA 90089-2564}

\cftpagenumbersoff{figure}
\cftpagenumbersoff{table}
\begin{document}
\maketitle

\begin{abstract}

A new stereoscopic image quality assessment database rendered using the
2D-image-plus-depth source, called MCL-3D, is described and the
performance benchmarking of several known 2D and 3D image quality
metrics using the MCL-3D database is presented in this work. Nine
image-plus-depth sources are first selected, and a depth image-based
rendering (DIBR) technique is used to render stereoscopic image pairs.
Distortions applied to either the texture image or the depth image
before stereoscopic image rendering include: Gaussian blur, additive
white noise, down-sampling blur, JPEG and JPEG-2000 (JP2K) compression
and transmission error. Furthermore, the distortion caused by imperfect
rendering is also examined. The MCL-3D database contains 693
stereoscopic image pairs, where one third of them are of resolution
$1024 \times 728$ and two thirds are of resolution $1920 \times 1080$.
The pair-wise comparison was adopted in the subjective test for user
friendliness, and the Mean Opinion Score (MOS) can be computed
accordingly. Finally, we evaluate the performance of several 2D and 3D
image quality metrics applied to MCL-3D. All texture images, depth
images, rendered image pairs in MCL-3D and their MOS values obtained in
the subjective test are available to the public
(http://mcl.usc.edu/mcl-3d-database/) for future research and
development.

\end{abstract}


{\noindent \footnotesize{\bf Address all correspondence to}: Hyunusk Ko, University of Southern
California, Ming Hsieh Department of Electrical Engineering, 3740 McClintock Avenue,
Los Angeles, CA, United States, CA 90089-2564; Tel: 213-740-4622; E-mail:  \linkable{kosu9980@gmail.com} }

\begin{spacing}{2}   

\section{Introduction}\label{sec:introduction}  

Stereoscopic image/video contents become popular nowadays.  Since the
multi-view image format \cite{Merkle2006} is costly for visual
communication, the 2D-image-plus-depth format \cite{Fehn2004} is
proposed as an alternative, where a texture image and its associated
depth image are recorded at a view point simultaneously. For
stereoscopic display, the depth image-based rendering (DIBR) technique
is applied to the texture and depth images to generate the proper left-
and right-views. The 2D-image-plus-depth format has a few advantages,
including bandwidth efficiency, interactivity and 2D/3D video content
switch, etc\cite{Solh2012}.  A 3D video coding standard, called MPEG-C
part 3\cite{Wkh2007}, has been developed using the Multi-View-plus-Depth
(MVD) format. In this work, we address the visual quality assessment
problem using the 2D-image-plus-depth source.  With the DIBR technology,
the stereoscopic images rendered and displayed on the stereoscopic
screen rely on the quality of texture images, depth maps and the
rendering technology.  Since discomfort caused by watching stereoscopic
images may go beyond annoying and lead to psychological dizziness, we
cannot over-emphasize the importance of the stereoscopic image/video
quality assessment problem.

\begin{figure}[htbp]
\begin{center}
\includegraphics[width=\columnwidth]{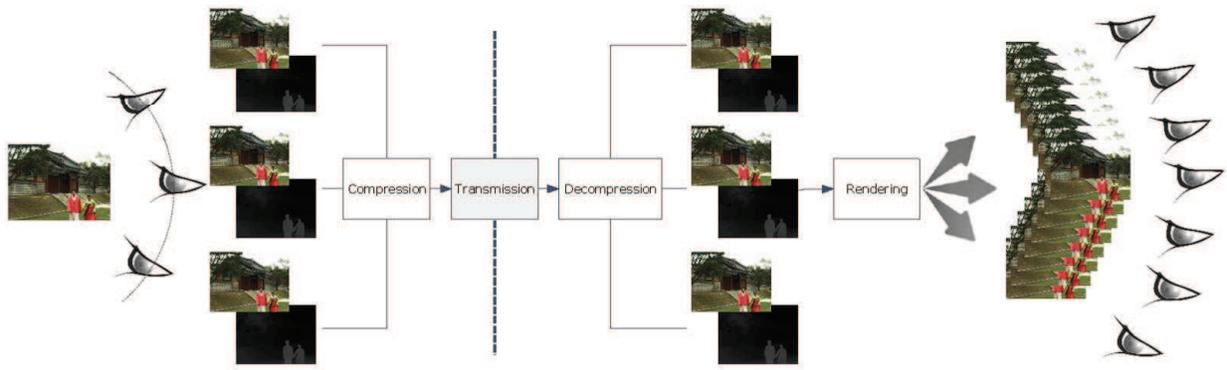}\\
\end{center}
\caption{The processing flow of a stereoscopic visual communication
system with the DIBR technology.}\label{fig:dibr-processing-flow}
\end{figure}

We show the processing flow of a stereoscopic visual communication
system with the DIBR technology in Fig.  \ref{fig:dibr-processing-flow}.
At the encoder end, the texture and depth images captured at one
viewpoint (or multiple viewpoints) are compressed and transmitted
separately. At the decoder end, texture and depth maps are decoded and a
pair of stereoscopic images can be rendered. In this work, we follow a
similar process to build a stereoscopic image quality assessment
database and consider a wide range of distortion types occurring in
video capturing, compression, transmission and rendering.  The resulting
database is called MCL-3D.

There are several publicly accessable stereoscopic image databases
developed for the quality assessment purpose as listed in Table
\ref{tab:summary-of-3d-database}.  Only symmetric distortions (i.e., the
same distortion type and level) are applied to the left and right images
in the LIVE Phase I database \cite{Moorthy2013}. Non-symmetric distortions
are considered in the LIVE Phase II database \cite{Chen2012} as a
generalization. The IVC 3D database \cite{Benoit2008} is similar to LIVE
Phase I yet with a different set of source images.  One common concern
with these three databases is that the resolution of stereoscopic images
is low. Images of higher resolution are adopted by the IVC DIBR
\cite{Bosc2011} and the EPFL databases \cite{Goldmann2010}.  One unique
feature of the EPFL database is that it examines the effect of different
disparity values on the resulting visual quality so as to develop a
guideline on disparity selection.  A similar yet more delicate work is
given in [\citenum{Voronov2013}], where the disparity effect on continuous
video is analyzed so that some visual metrics can be fine-tuned for
disparity selection in 3D films. The IVC DIBR database examined the
visual quality of rendered stereoscopic pairs with various rendering
mechanisms. However, no transmission distortion is considered.
Furthermore, distortions were imposed on binocular images directly,
which has a more restricted application constraint.

\begin{table}[htbp]\small
\centering
\caption{Summary of 3D Image Databases}\label{tab:summary-of-3d-database}
\begin{threeparttable}
\begin{tabular}{p{1.8cm}p{1.8cm}p{1.8cm}p{1.8cm}p{1.8cm}p{1.8cm}} \hline
          & LIVE\_I & LIVE\_II & IVC 3D & IVC DIBR Image & EPFL \\ \hline
Scenes & 20    & 8     & 6     & 3     & 10 \\
Image Resolution  & 640x360 & 640x360 & 512x448\tnote{1} & 1024x768 & 1920x1080 \\
\multicolumn{ 1}{r}{Distortion types} & Blur,  & White noise,  & Blur, & Holl filling & Disparity \\
\multicolumn{ 1}{r}{} & Fast fading,  & Blur, & JPEG,  &       &  \\
\multicolumn{ 1}{r}{} & JP2K,  & JPEG,  & JP2K  &       &  \\
\multicolumn{ 1}{r}{} & JPEG,  & JP2K, &       &       &  \\
\multicolumn{ 1}{r}{} & White noise & Fast fading &       &       &  \\
Distortion Levels & -\tnote{2}     & -\tnote{3}    & -\tnote{4}     & 7\tnote{5}     & 10\tnote{6} \\
Total Num & 385   & 368   & 96    & 96    & 100 \\  \hline
\end{tabular}
  \begin{tablenotes}
    \footnotesize
    \item[1] Image size is not identical in IVC 3D, 512x448 is the mean value provided in the corresponding paper.
    \item[2] Different distortion types have different levels in LIVE Phase I.
    \item[3] LIVE Phase II has complex level definitions for asymmetrical distortion types.
    \item[4] Different distortion types have different levels in IVC 3D database.
    \item[5] IVC DIBR database has 7 different hole filling algorithms, taken as 7 distortion levels.
    \item[6] 10 camera configurations, taken as 10 distortion levels.
\end{tablenotes}
\end{threeparttable}
\end{table}

In contrast, distortions are applied to either the texture image or the
depth image before stereoscopic image rendering in MCL-3D.  The
distortion types of consideration include: Gaussian blur, additive white
noise, down-sampling blur, JPEG and JPEG-2000 (JP2K) compression and
transmission error.  The artifact caused by imperfect rendering is also
considered.  The pair-wise comparison was adopted in the subjective test
to be friendly to viewers, and the Mean Opinion Score (MOS) was computed
accordingly. All texture images, depth images, rendered image pairs and
their MOS values obtained from the subjective test in MCL-3D are
available to the public (http://mcl.usc.edu/mcl-3d-database/)
for future research and development.

The rest of this paper is organized as follows. The source data, the
DIBR rendering process and distortions adopted by the MCL-3D database
are detailed in Sec. \ref{sec:description}.  The human subject test
process is presented in Sec.  \ref{sec:subjective}. Then, we compare
several existing 2D and 3D objective image quality assessment methods
against the MCL-3D database in Sec. \ref{sec:test}. Finally, concluding
remarks and future work are given in Sec.  \ref{sec:conclusion}.

\section{Description of MCL-3D Database}\label{sec:description}
\subsection{Stereoscopic Image Pair Synthesis System}

\begin{figure}[htbp]
\begin{center}
\includegraphics[width=0.5\columnwidth]{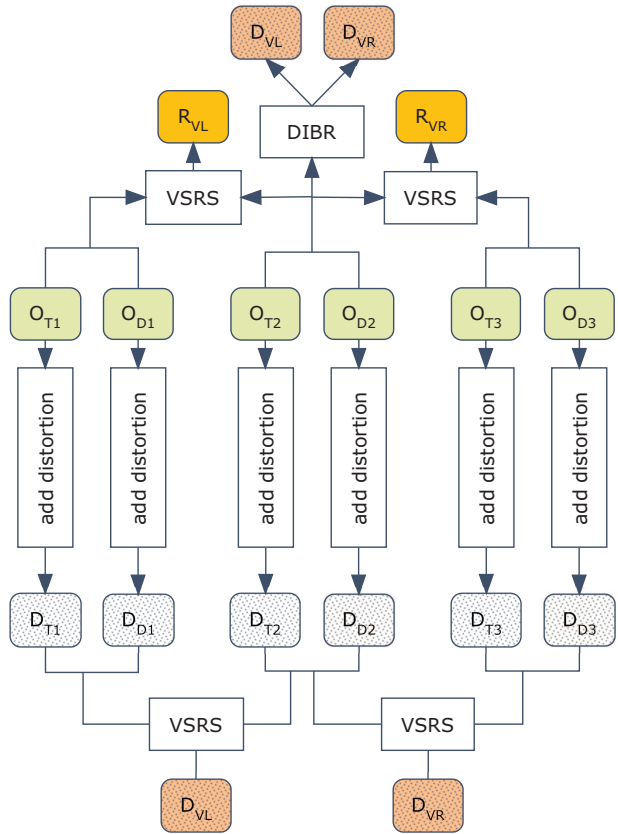}
\end{center}
\caption{The block-diagram of the stereoscopic image pair synthesis
system used to create the MCL-3D database.}
\label{fig:structure-of-experiment}
\end{figure}

The stereoscopic image pair synthesis system used to create the MCL-3D
database is shown in Fig. \ref{fig:structure-of-experiment}, where
characters $O$, $D$ and $R$ denote original input, distorted and
rendered outputs, and subscript characters $T$, $D$ and $VL$ and $VR$
denote the texture image, depth map, rendered left-view and right-view,
respectively.  First, the original texture image and its associated
depth map of three views, denoted by ($O_{T1}$, $O_{D1}$), ($O_{T2}$,
$O_{D2}$), and ($O_{T3}$, $O_{D3}$), are obtained by selecting key
frames from 3DVC test sequences \cite{ISO/IECJTC1/SC29/WG112011a}
 and used as the input.  Distortions of different types
and levels were introduced to either the texture image or the depth map,
and distorted texture images or depth maps are used as the input to the
view synthesis reference software (VSRS) \cite{Wg11.sc29.org2010} to
render the distorted stereoscopic image pair. For the DIBR distortion,
we take the original source $O_{T2}$ and $O_{D2}$ as the input, and use
four different rendering algorithms to generate the stereoscopic image
pair.  The VSRS offers a near-prefect stereoscopic image synthesis
mechanism.  If the original left- and right-views are given, the VSRS
can output a near perfect rendered view in between.  The rendered
left-view and right-view using the original texture images and depth
maps, denoted by $R_{VL}$ and $R_{VR}$, will be taken as the reference
for further analysis.

\subsection{Image and Depth Source}

\begin{figure}[htbp]
\begin{center}
\includegraphics[width=0.95\columnwidth]{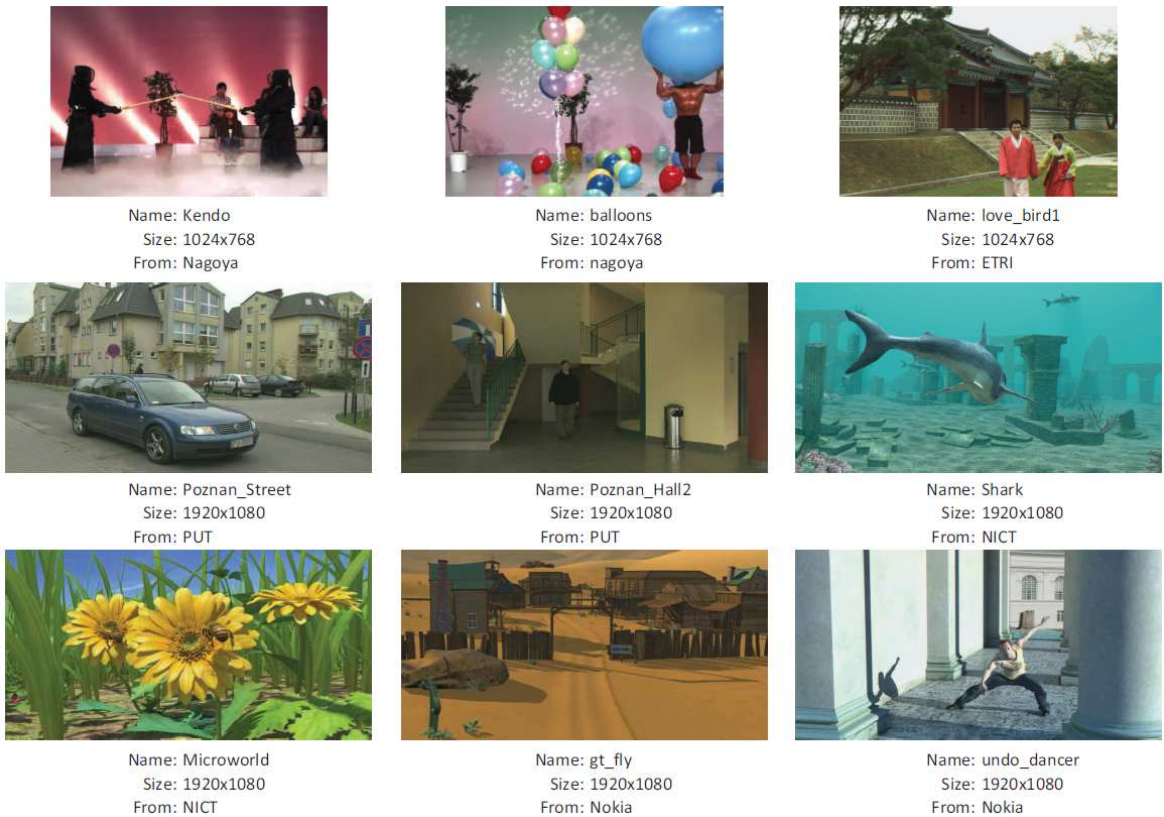}\\
\end{center}
\caption{Reference images in the MCL-3D database.}\label{fig:source-images}
\end{figure}

The quality of a database is highly dependent on reference images. The
selected images should be representative and with sufficient diversity.
The test sequences used in the 3DVC standard can be good candidates,
which provide a few multi-view sequences associated with depth maps.  We
removed those of uncommon spatial resolution and/or with a camera
calibration problem from this candidate set and, finally, selected nine of
them as the reference images in the MCL-3D database. They are shown in
Fig. \ref{fig:source-images}.

\subsection{Distortion Types and Levels}

In a communication system that adopts the 3DVC coding standard,
distortions may come from various stages such as image acquisition,
compression, transmission and rendering. Gaussian blur and additive
noise may occur in the acquisition stage. The image and the depth map
may be down-sampled to accommodate multiple display devices before
compression. For efficient transmission, all images should be
compressed, which leads to blockiness and compression blur. Transmission
errors may occur in the transmission stage. A rendering algorithm will
be adopted to render multiple views for display. Some of these
distortions were investigated before as shown in Table
\ref{tab:summary-of-3d-database}.  We include distortions of all
above-mentioned cases in the MCL-3D database.

\begin{figure}[htbp]
\begin{center}
\includegraphics[width=\columnwidth]{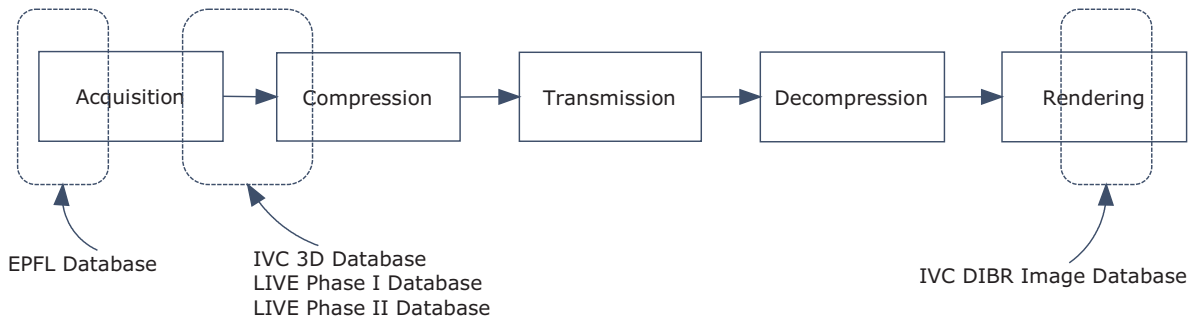}\\
\end{center}
\caption{The entire precessing flow and the corresponding distorted part
for each database} \label{fig:entire-processing-flow}
\end{figure}

Based on the recommendations of ITU \cite{ITU-T, ITU-T2000, ITU-R2002}
and VQEG \cite{VQEG, VQEGa}, we consider five quality levels in
subjective tests. The original reference stereoscopic images have the
``excellent'' quality while the other 4-level distorted images
correspond to ``very good'', ``good'', ``fair'' and ``poor'',
respectively. The distortion caused by imperfect rendering has not been
well studied before. Typically, only the mid-view image and its depth
map are taken as the input, and a stereoscopic image pair is rendered
using a hole filling technique. In our experiment, we take $O_{T2}$ and
$O_{D2}$ as the input to generate the stereoscopic image pair.
Distortion types are summarized in Table \ref{tab:level-detail} and
explained below.

\begin{table}[htbp]\small
\centering
\caption{Distortion generation mechansims and the associated level
parameters.}\label{tab:level-detail}
\includegraphics[width=\columnwidth]{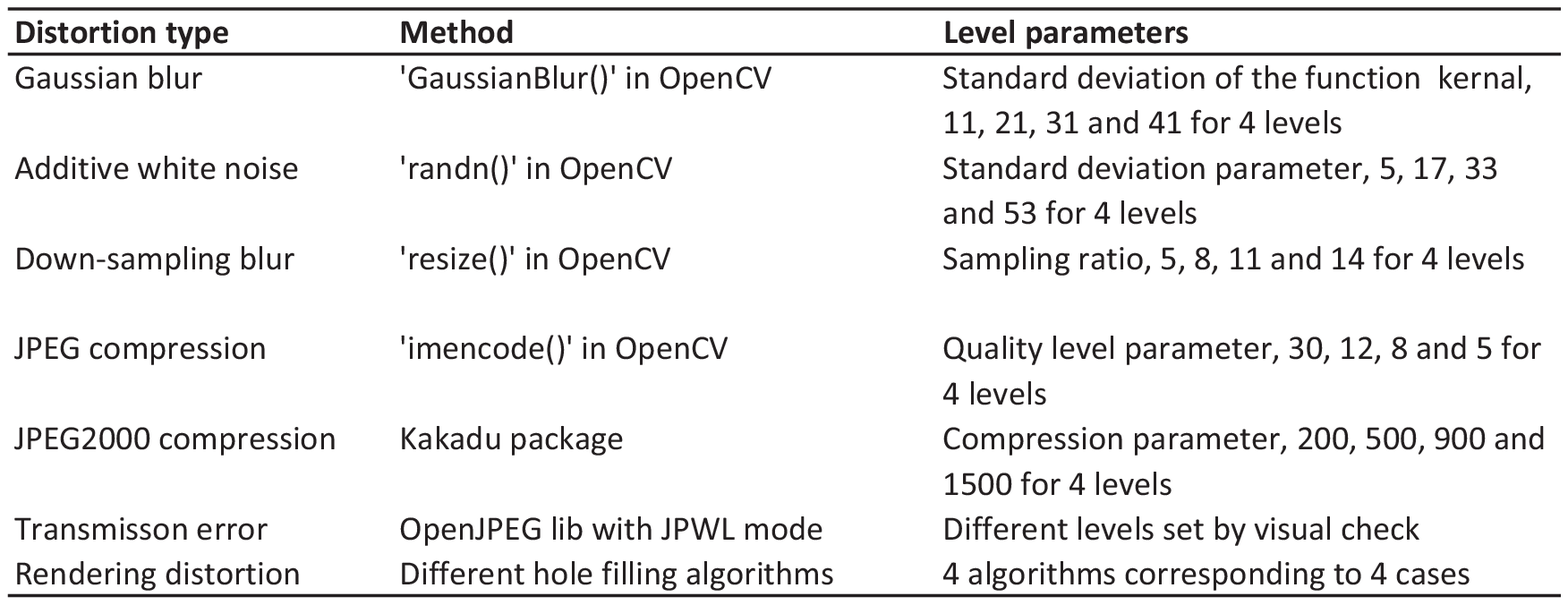}\\
\end{table}

\begin{itemize}
\item Gaussian Blur\\
Many parameters have to be calibrated \cite{Goldmann2010} during the
acquisition of high quality stereoscopic images, wherein the focal
length is a critical one. Texture images from any view will be blurred
due to an improper focal length. Depth maps could be either acquired by
equipments\cite{Izadi:2011:KRR:2047196.2047270, Du:2011:IMI:2030112.2030123}
or estimated by depth estimation
algorithms\cite{M.TanimotoT.Fujiia}.  It was claimed by some
researchers\cite{Fehn2004, Bosc2011} that the visual experience
can be improved by applying some blur to the depth map before rendering.
Its effectiveness can be studied using MCL-3D. We used `GaussianBlur()'
function in the OpenCV\cite{opencv_library} library to add the Gaussian blur
effect and controlled distortion levels by varying the standard deviation
parameter of the kernel. Their values were set to 11, 21, 31, 41 for four
distortion levels.

\item Additive White Noise\\
In digital image capturing systems, CMOS or CCD sensors are used to
capture R/G/B color light intensities. The intensity is later
transformed to the voltage and quantized to digital pixel values.
Interference is ubiquitous in electronic circuits. It appears in
form of additive white noise in the texture or depth image. The
`randn()' function in the OpenCV library was used to generate additive
noise whose levels were controlled by selecting four standard deviation
values (5, 17, 33 and 53).

\item Down-sampling Blur\\
The captured image may be down-sampled to fit a different spatial
resolution requirement. The `resize()' function in OpenCV is used for
down-sampling and up-sampling. Four different down-sampling blur levels
with a sampling ratio of 5, 8, 11 and 14 were included.

\item JPEG and JP2K Compression\\
We applied JPEG and JP2K compression to source images. For JPEG
compression, we utilized the `imencode()' function in OpenCV with four
quality levels (30, 12, 8 and 5). For JP2K compression, we utilized the
Kakadu\cite{NewSouthInnovationsPtyLtd} package with four compression
parameters (200, 500, 900 and 1500) for four distortion levels.

\item Transmission Error\\
We used the OpenJPEG library to encode source images and then applied
unequal protection and error correction codes in the JPWL mode.  Some
bit errors were added to the compressed bitstreams. At the decoder side,
the errors were partly corrected. With the assistance of protection
methods, it is difficult to build a simple relationship between the
bit-error rate and the visual quality of the decoded image. Thus,
we used 80 seeds to generate a group of error-corrupted images and
selected 4 from them to obtain 4 transmission error levels.

\item Rendering Distortion\\
Stereoscopic images were rendered based on the texture and the depth map
images using the DIBR technology.  Typical rendering errors include the
black hole\cite{Tian2009} and the boundary blur, which tend to appear
with imperfect rendering techniques \cite{Bosc2011}. We selected several
representative ones, including DIBR without hole filling, DIBR with
filtering\cite{Fehn2004}, DIBR with inpainting\cite{Telea2004}, and DIBR
with hierarchical hole filling\cite{Solh2012, Solh2012a, Solh2010}.
\end{itemize}

\section{Subjective Test}\label{sec:subjective}

For the subjective test, the test environment was set up according to
the ITU recommendations\cite{ITU-T} and a pairwise comparison method was
adopted. Testing results were verified after the subjective test
procedure.

ITU and VQEG are two organizations working on the standardization of
subjective test methods. Both of them have published recommendations on
subjective test procedure for 2D images\cite{ITU-R2002}, 2D
videos\cite{ITU-T} and stereoscopic images\cite{ITU-T2000}. They can be
roughly classified into four groups according to score levels and
stimulus numbers as shown in Table \ref{tab:subjective-test-methods}.

\begin{table}[htbp]
  \centering
  \caption{Recommendations for subjective test methods}
  \begin{threeparttable}
    \begin{tabular}{l|l|l}
    \hline
    \textbf{} & \textbf{Discrete Score} & \textbf{Continuous Score} \\
    \hline
    \textbf{Single Stimulus} & ACR\tnote{1}, ACR-HR\tnote{2} & SSCQE\tnote{3} \\
    \hline
    \textbf{Double Stimulus} & DCR\tnote{4}, DSIS\tnote{5}, CCR\tnote{6}, DSCS\tnote{7} & DSCQS\tnote{8}, SAMVIQ\tnote{9} \\
    \hline
    \end{tabular}%
  \label{tab:subjective-test-methods}%

  \begin{tablenotes}
    \footnotesize
    \item[1] ACR: Absolute Category Rating.
    \item[2] ACR-HR: Absolute Category Rating with Hidden Reference.
    \item[3] SSCQE: Single stimulus continuous quality evaluation.
    \item[4] DCR: Degradation category rating.
    \item[5] DSIS: Double Stimulus Impairment Scale.
    \item[6] CCR: Comparison Category Rating.
    \item[7] DSCS: Double Stimulus Comparison Scale.
    \item[8] DSCQS: Double Stimulus Continuous Quality Scale.
    \item[9] SAMVIQ: Subjective Assessment Methodology for Video Quality.
\end{tablenotes}
  \end{threeparttable}
\end{table}%

It was mentioned in [\citenum{Tominaga2010}] and
[\citenum{BROTHERTON2006}] that the continuous scale score does not
improve the precision of test results.  For the ACR method, the same
score may have a different meaning for a different assessor.  Even for
the same assessor, the rating criteria may vary along test time. For
this reason, we focus on methods with double stimulus and discrete
scores.

Furthermore, we adopted the pairwise comparison method in the MCL-3D
database. The pairwise comparison method has solid mathematical
foundation and is extensively used for resource ranking and
recommendation systems.  Generally speaking, two stereoscopic image
pairs are viewed by an assessor simultaneously and, then, the assessor
selects the preferred one so as to assign a point score. The point score
of a stereo image pair will accumulate across multiple rounds of
pairwise competition, and the final point score is properly normalized
to yield the final opinion score for the same assessor. The opinion
scores of multiple assessors are averaged to result in the final mean
opinon score (MOS) for each stereoscopic image pair.

\begin{figure}[htbp]
  \begin{center}
  \includegraphics[width=0.6\columnwidth]{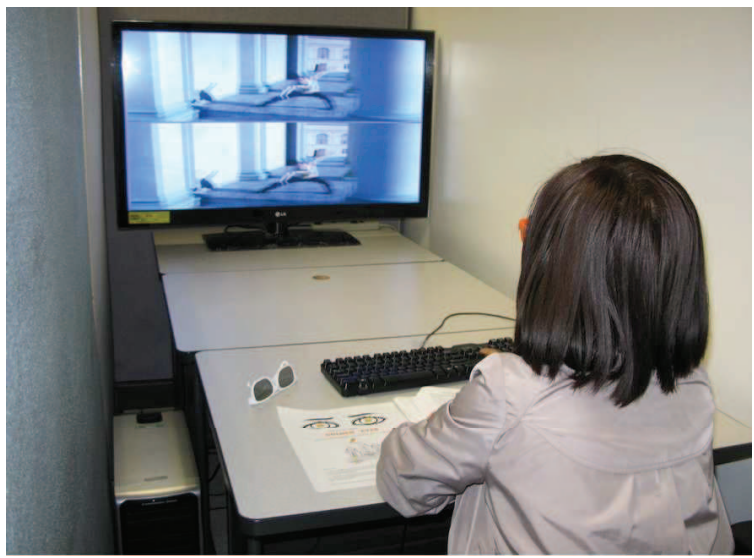}\\
  \end{center}
  \caption{Illustration of the subjective test environment.}
  \label{fig:subjective-test-environment}
\end{figure}

The subjective test environment is described below. The display
equipment was 46.9'' LG 47LW5600. Assessors were seated 3.2 meters away
from the display screen as shown in Fig.
\ref{fig:subjective-test-environment} \cite{ITU-T2000}. During the test,
two stereoscopic image pairs were shown on the screen simultaneously.
The images were resized to adapt to the display, and the gap between two
images was padded with grey levels as specified in
[\citenum{ITU-T2000}].  With pair-wise comparison, only the relative
quality of the two pairs was annotated by the assessor and the resize
operation had little affect on the final result.

\begin{figure}[htbp]
  \begin{center}
  \includegraphics[width=0.6\columnwidth]{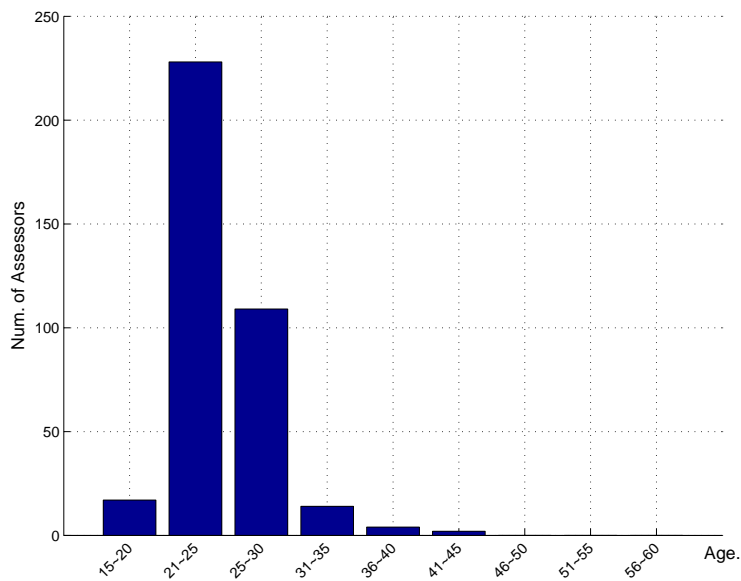}\\
  \end{center}
  \caption{The age distribution of assessors.}
  \label{fig:age-distribution}
\end{figure}

We developed a program with a proper GUI interface to control the
quality assessment process for each assessor. For each image set, the
test time ranged from 12 to 15 minutes so as to comply with the
recommendation in ITU-R Rec. BT.500\cite{ITU-R2002}.  After the
subjective test, we conducted a short interview with the assessor for
their evaluation experience. The assessors were students from the
University of Southern California in USA. Among the 270 assessors,
there were 170 males (63\%) and 100 females (37\%).  In order to
investigate the score difference between experts and non-experts,
we asked assessors about their familiarity on stereoscopic images.
Among them, 34 (or 13\%) were experts and 236 (or 87\%) were non-experts.
The age distribution of the assessors is given in Fig. \ref{fig:age-distribution}.

Each assessor conducted the evaluation of all distorted images for one
reference image in one test session.  We collected 30 opinion scores for
every distorted image pair.  The subjective test results were further
filtered by a screening process \cite{ITU-R2002}.  In building the
MCL-3D database, the highest 10\% and the lowest 10\% scores for each
image were treated as outliers and discarded.  The final MOS was
calculated as the mean of remaining 24 opinion scores.  The recommended
number of assessors is 15 by ITU \cite{ITU-R2002} and 24 by
VQEG\cite{Bosc2011} for images.  Bosc\cite{Bosc2011} tested the number
of assessors for the subjective test for synthesized 3D view and
concluded that the minimum number is 32 for ACR and less than 24 for
pairwise comparison. Thus, our MOS calculation does meet the
requirements of all above recommendations.

A summary of the MCL-3D database is given in Table \ref{tab:summary-of-MCL-3D}.

\begin{table}[htbp]
  \centering
  \caption{Summary of MCL-3D database}
    \begin{tabular}{rl}
    \hline
    \textbf{Main Characters} & \textbf{MCL\_3D database} \\
    \hline
    Scenes & 9 \\
    \multicolumn{ 1}{r}{Image resolution } & 6 with 1920x1080 \\
    \multicolumn{ 1}{r}{} & 3 with 1024x768 \\
    \multicolumn{ 1}{r}{Distortion types} & Gaussian blur,  \\
    \multicolumn{ 1}{r}{} & Down-sampling blur, \\
    \multicolumn{ 1}{r}{} & Additive white noise \\
    \multicolumn{ 1}{r}{} & JPEG compression, \\
    \multicolumn{ 1}{r}{} & JP2K compression, \\
    \multicolumn{ 1}{r}{} & Transmission error, \\
    \multicolumn{ 1}{r}{} & Rendering algotirhm \\
    Distortion levels & 4 \\
    Total num of image pairs & 693 \\
    Subjective test method & pair-wise comparison \\
    No. of assessors & 270 \\
    Scale of MOS & 0...9 \\
    \hline
    \end{tabular}%
  \label{tab:summary-of-MCL-3D}%
\end{table}%

\section{Performance Comparison of Objective Quality Indices}\label{sec:test}

In this section, we compare the performance of several objective quality
indices against the MCL-3D database.

\subsection{Performance of 2D IQA Indices}

There are quite a few 2D image quality assessment methods proposed in
literature \cite{Lin2011}. Traditionally, image distortion indices focus
on fidelity by measuring the exact difference between the distorted and
the reference images; {\em e.g.} the mean-squared error (MSE), the Peak
Signal to Noise Ratio (PSNR), etc. The fidelity concept has been
scrutinized and challenged by researchers recently.  New image quality
indices were proposed. Examples include the Noise Quality Measure (NQM)
\cite{Damera-Venkata2000}, the Universal Quality Index (UQI)
\cite{Wang2002}, the Structural Similarity Index (SSIM) \cite{Wang2004},
the Multi-scale Similarity Index (MS-SSIM) \cite{Wang}, the Feature
Similarity Index \cite{Zhang2011}, the visual information fidelity (VIF)
\cite{Sheikh2006}, the pixel-based VIF (VIFP) \cite{Sheikh2006}, the
visual signal-to-noise ratio (VSNR) \cite{Chandler2007}, the image
fidelity criterion (IFC) \cite{Sheikh2005}, PSNR-HVS
\cite{Ponomarenko2007} and C4 \cite{Carnec2003}.

We applied these quality indices to the left- and the right-views of the
stereoscopic image pairs and obtained their mean as the quality score.
We conducted this test on the MCL-3D database as well as two other
stereoscopic image databases; namely, LIVE Phase I \cite{Zhang2011} and
IVC 3D \cite{Benoit2008}. The Pearson Correlation Coefficient (PCC), the
Spearman rank order correlation coefficient (SROCC) and the
mean-squared-error (MSE) between the MOS and the objective scores are
shown in Table \ref{tab:2D-benchmarks}. We see that both the PCC and
SROCC values of these indices are less than 90\% against MCL-3D. There
is certainly room for further improvement.

{\small
\begin{table}[htbp]
\centering
\caption{Performance comparison of 2D objective quality indices applied to
MCL-3D, LIVE Phase I and IVC databases.}\label{tab:2D-benchmarks}
\begin{tabular}{llllllllll}
\hline
\multicolumn{ 1}{r}{Indices} & \multicolumn{ 3}{c}{MCL-3D} & \multicolumn{ 3}{c}{LIVE Phase I} & \multicolumn{ 3}{c}{IVC} \\
\multicolumn{ 1}{r}{} & PCC  & SROCC & MSE   & PCC  & SROCC & MSE   & PCC  & SROCC & MSE \\
\hline
C4    & 0.8683 & 0.8690 & 0.6452 & 0.9078 & 0.9144 & 0.0596 & 0.7874 & 0.7304 & 0.1700 \\
IFC   & 0.7395 & 0.7398 & 0.8757 & 0.5466 & 0.9071 & 2.7000 & 0.7051 & 0.6135 & 0.1955 \\
MS\_SSIM & 0.8656 & 0.8763 & 0.6514 & 0.7382 & 0.6093 & 0.0972 & 0.7676 & 0.6919 & 0.1767 \\
NQM   & 0.8684 & 0.8694 & 0.6451 & 0.8349 & 0.8461 & 3.3030 & 0.6816 & 0.5973 & 0.2018 \\
PSNR\_HVS & 0.8783 & 0.8857 & 0.6220 & 0.7563 & 0.8042 & 3.4906 & 0.7089 & 0.6374 & 0.1945 \\
PSNR  & 0.8320 & 0.8405 & 0.7218 & 0.6482 & 0.6529 & 3.9963 & 0.5843 & 0.5554 & 0.2238 \\
SSIM  & 0.7654 & 0.7834 & 0.8372 & 0.6977 & 0.6616 & 0.1391 & 0.6817 & 0.6478 & 0.2017 \\
UQI   & 0.7372 & 0.7551 & 0.8789 & 0.9007 & 0.8974 & 0.0750 & 0.5706 & 0.5244 & 0.2265 \\
VIFP  & 0.7770 & 0.7897 & 0.8188 & 0.8266 & 0.8681 & 0.0660 & 0.7355 & 0.6869 & 0.1868 \\
VIF   & 0.7762 & 0.7929 & 0.8202 & 0.8883 & 0.9002 & 0.0664 & 0.7971 & 0.7083 & 0.1665 \\
VSNR  & 0.8289 & 0.8370 & 0.7277 & 0.7317 & 0.7847 & 5.0763 & 0.6723 & 0.6110 & 0.2041 \\
\hline
\end{tabular}%
\end{table}%
}

\subsection{Performance of 3D IQA Indices}

Several IQA indices have been developed to target at stereoscopic image
pairs. Campisi\cite{Campisi2007} conducted a preliminary test on the
acuity difference between different eyes and found no apparent
difference.  Ryu\cite{Ryu2012} proposed an extended version of the SSIM
index based on a binocular model. Their index uses a fixed set of
parameters and is not adaptive to asymmetric distortions.
Ko\cite{Ko2013} introduced the structural distortion parameter (SDP),
which varies according to different distortion types. The SDP was
employed as a control parameter in a binocular perception model to
provide robust QA results for both symmetric and asymmetric distortions.
Gorley\cite{Gorley2008} used the difference of relative contrast between
the reference image pair and distorted image pair to derive the quality
index. Benoit\cite{Benoit2008} extracted the disparity maps from both
the reference and the distorted image pairs, calculated the distortion
between them, and integrated it with other factors to form the final
quality index. Sazzad\cite{Sazzad2012} exploited the disparity map and
performed several integration methods to derive the quality index.

We evaluated the following four indices against the MCL-3D, the LIVE
Phase I, and the IVC databases:
\begin{itemize}
\item Method A \cite{Ryu2012},
\item Method B \cite{Benoit2008},
\item Method C \cite{Campisi2007},
\item Method D \cite{Ko2013}.
\end{itemize}
The PCC, SROCC and MSE results are shown in Table \ref{tab:3D-benchmarks}.
We see that these 3D IQA indices do not show much superiority over 2D IQA
indices. How to derive a better 3D IQA index is still a challenging
problem.

{\small
\begin{table}[htbp]
\centering
\caption{Benchmarks of 3D quality assessment metrics}
\label{tab:3D-benchmarks}%
    \begin{tabular}{llllllllll}
    \hline
    \multicolumn{ 1}{r}{Metric} & \multicolumn{ 3}{c}{MCL-3D} & \multicolumn{ 3}{c}{LIVE Phase I} & \multicolumn{ 3}{c}{IVC} \\
    \multicolumn{ 1}{r}{} & PCC  & SROCC & MSE   & PCC  & SROCC & MSE   & PCC  & SROCC & MSE \\
    \hline
    Method A & 0.8419 & 0.8503 & 0.7020 & 0.6775 & 0.6075 & 0.1286 & 0.7579 & 0.6869 & 0.1799 \\
    Method B & 0.7545 & 0.7672 & 0.8537 & 0.8174 & 0.8493 & 0.0930 & 0.2851 & 0.4916 & 0.2643 \\
    Method C & 0.8683 & 0.8690 & 0.6452 & 0.9067 & 0.9133 & 0.0600 & 0.7873 & 0.7295 & 0.1700 \\
    Method D & 0.8910 & 0.8880 & 0.6055 & 0.9080 & 0.9050 & 6.8870 & 0.8410 & 0.8030 & 11.1200 \\
    \hline
    \end{tabular}%
\end{table}%
}

\section{Conclusion and Future Work}\label{sec:conclusion}

In this work, a detailed description of a stereoscopic image quality
assessment database called MCL-3D was given, and the performance
benchmarking of several known 2D and 3D image quality metrics using the
MCL-3D database were presented.  Distortions applied to the texture
image or the depth image before stereoscopic image rendering include:
Gaussian blur, additive white noise, down-sampling blur, JPEG and
JPEG-2000 (JP2K) compression and transmission error.  Furthermore, we
evaluated the performance of several 2D and 3D image quality metrics
applied to MCL-3D.  The MCL-3D database is available to the public for
future research and development.

Based on the experimental results, we see that none of the existing
objective quality metrics can provide satisfactory performance for
several stereoscopic image quality databases, including MCL-3D. It is
still an open problem to design a good objective quality method for 3D
imges. The learning-based methodology in [\citenum{Liu2013}] and
[\citenum{Liu2011b}] offers an effective solution to the 2D image
quality assessment problem.  We currently focus on the design of a good
stereoscopic image quality metric along the same direction.

\acknowledgments
This project is funded by the China Scholarship Council and the Samsung
Advanced Institute of Technology. The author would like to thank Mathieu
Carnec \cite{Carnec2003}, Mashhour M.  Solh \cite{Solh2012, Solh2012a, Solh2010}
and Seungchul Ryu \cite{Ryu2012} for providing their codes to test the MCL-3D database.
Computation for the work described in this paper was supported by the
University of Southern California Center for High-Performance Computing and
Communications.


\bibliography{library_without_url}   
\bibliographystyle{spiejour}   


\vspace{2ex}\noindent{\bf Rui Song} received his Bachelor degree in
Telecommunications, Master and Ph.D. degrees in Signal and Information
Processing from Xidian University, Xi'an, China, in 2003, 2006 and 2009,
respectively. From 2013 to 2014, he is a Postdoctoral Scholar in Professor
C.-C. Jay Kuo's Group, Ming Hsieh Department, Viterbi School of Electric
Engineering, University of Southern California, USA. Since 2010, Dr. Rui Song
has been an Associate Professor in School of Telecommunication Engineering
at Xidian University, Xian, China. He is now a member of Image and Video
Processing Lab in State Key Laboratory of Integrated Service Networks, and is
responsible for video group. His research interests include pre- and post-processing
of high definition camera, image and video quality assessment, video coding
algorithms, VLSI architecture design for image and video processing,
architecture design of video codec IP. He has published 20 research papers
which are indexed by SCI or EI and has got 4 patents. For more detailed
information about Professor Rui Song, please
refer to \linkable{http://web.xidian.edu.cn/songrui/en/index.html}.

\vspace{2ex}\noindent{\bf Hyunsuk Ko} received the B.S. and M.S. degrees
from the Department of  Electrical Engineering, Yonsei University, Seoul,
Korea, in 2006 and 2009, respectively. From 2009 to 2010, he was a
software engineer at Samsung Electronics, Suwon, Korea.
He is currently working toward the Ph.D. degree in the Department of
Electrical Engineering, University of Southern California, Los Angeles.
His research interests include 3D image/video processing, quality
assessment, big data analysis, and machine learning.

\vspace{2ex}\noindent{\bf Dr. C.-C. Jay Kuo} received the B.S. degree from
the National Taiwan University, Taipei, in 1980 and the M.S. and Ph.D.
degrees from the Massachusetts Institute of Technology, Cambridge, in 1985
and 1987, respectively, all in Electrical Engineering. From October 1987 to
December 1988, he was Computational and Applied Mathematics Research
Assistant Professor in the Department of Mathematics at the University of
California, Los Angeles. Since January 1989, he has been with the University
of Southern California (USC). He is presently Director of the Multimedia
Communication Lab. and Professor of Electrical Engineering and Computer
Science at USC.

His research interests are in the areas of multimedia data compression,
communication and networking, multimedia content analysis and modeling, and
information forensics and security. Dr. Kuo has guided 118 students to their
Ph.D. degrees and supervised 23 postdoctoral research fellows. Currently, his
research group at USC has around 30 Ph.D. students, which is one of the
largest academic research groups in multimedia technologies. He is co-author
of about 220 journal papers, 850 conference papers and 12 books. He delivered
around 550 invited lectures in conferences, research institutes, universities
and companies. He ranks as the top advisor in the Mathematics Genealogy
Project in terms of the number of supervised PhD students.
Dr. Kuo is a Fellow of AAAS, IEEE and SPIE. He is Editor-in-Chief for the
IEEE Transactions on Information Forensics and Security and Editor Emeritus
for the Journal of Visual Communication and Image Representation (an Elsevier
journal). He was Editor-in-Chief for the Journal of Visual Communication and
Image Representation in 1997-2011. He was on the Editorial Board of the IEEE
Signal Processing Magazine in 2003-2004, IEEE Transactions on Speech and
Audio Processing in 2001-2003, IEEE Transactions on Image Processing in
1995-98 and IEEE Transactions on Circuits and Systems for Video Technology in
1995-1997.

Dr. Kuo received the National Science Foundation Young Investigator Award
(NYI) and Presidential Faculty Fellow (PFF) Award in 1992 and 1993,
respectively. He received the Northrop Junior Faculty Research Award from the
USC Viterbi School of Engineering in 1994. He received the best paper awards
from the Multimedia Communication Technical Committee of the IEEE
Communication Society in 2005, from the IEEE Vehicular Technology Fall
Conference (VTC-Fall) in 2006, and from IEEE Conference on Intelligent
Information Hiding and Multimedia Signal Processing (IIH-MSP) in 2006. He was
an IEEE Signal Processing Society Distinguished Lecturer in 2006, a recipient
of the Okawa Foundation Research Award in 2007, the recipient of the
Electronic Imaging Scientist of the Year Award in 2010, the holder of the
Fulbright-Nokia Distinguished Chair in Information and Communications
Technologies from 2010-2011, and a recipient of the Pan Wen-Yuan Outstanding
Research Award in 2011. He is President of Asia Pacific Signal and Information
Processing Association (APSIPA) from 2012 to 2014.


\listoffigures
\listoftables

\end{spacing}
\end{document}